\documentclass[letterpaper]{article}
\usepackage[preprint]{aaai2027}
\usepackage[hyphens]{url}
\usepackage{graphicx}
\usepackage{natbib}
\usepackage{caption}
\usepackage{algorithm}
\usepackage{algorithmic}
\usepackage{booktabs}
\usepackage{amsmath}
\usepackage{amssymb}
\usepackage{array}
\usepackage{placeins}

\newcommand{\carve}{\textsc{CARVE}}
\newcommand{\ego}{\mathrm{ego}}
\newcommand{\bj}{B_j}
\newcommand{\amax}{\alpha^{\max}_j}
\newcommand{\cert}{\mathcal{C}}
\newcommand{\repair}{\mathcal{A}}
\newcommand{\fallback}{\mathcal{A}_{\mathrm{fb}}}
\newcommand{\rules}{\mathcal{H}}
\newcommand{\ops}{\mathcal{O}}
\newcommand{\pool}{\mathcal{P}}
\newcommand{\phiobj}{\Phi}
\newcommand{\cost}{\rho}
\newcommand{\gain}{\Delta g}
\newcommand{\Req}{\Delta}
\newtheorem{theorem}{Theorem}

\title{CARVE: Certified Affordable Repair of Vetoed Maneuvers via Envelopes for Interactive Driving}
\author{Yifan Wang}
\affiliations{
    Department of Mechanical Engineering, McGill University, QC, H3A 2T7, Canada\\
    yifan.wang18@mail.mcgill.ca
}

\begin{document}

\maketitle

\begin{abstract}
Interactive driving exposes a failure mode that is easy to miss in rule-aware
autonomous-driving stacks: a hard-rule margin can be negative for an ego
candidate even though a small lawful accommodation by a non-priority agent
would restore feasibility. Existing rulebooks, shields, and reachability
filters are strong at vetoing unsafe actions, while prediction-based planners
model likely responses. Neither returns a runtime proof object that states
which bounded multi-agent edit repairs the maneuver, who owns the edit, whether
the request is right-of-way affordable, and what ego fallback remains if the
request is not observed. We formulate this missing object as \emph{interactive
repair certification} and introduce \carve, a prediction-free certificate layer
over a finite lattice of ego-owned and agent-owned tactical operators.
Agent-owned requests are admissible only inside
\(\bj(s)=\beta(\pi_j)\amax(s)\), a cooperation envelope that separates
kinematic reachability from normative priority. The resulting certificate
records the binding rule, repair category, repair set,
responsibility-weighted cost split, and fallback. On 589
Lanelet2-geometry-grounded INTERACTION replay episodes, \carve-Greedy accepts
98.64\% of initially vetoed maneuvers and recovers 370/378 human-resolved false
vetoes, while preserving 589/589 right-of-way respect, zero priority-agent
false positives, and 400/400 negative-stress vetoes. We prove certificate
soundness, structural right-of-way respect, exact finite-lattice minimality,
fallback contingency, and blame-consistency conditions. \carve\ does not
predict or require another driver's compliance; it certifies whether a proposed
interaction is bounded, attributable, and normatively admissible under declared
assumptions.
\end{abstract}

\section{Introduction}

Autonomous-driving systems have made rapid progress in perception, forecasting,
and motion generation, but their deployment still depends on a more basic
property: the vehicle must act safely, legally, and transparently in interactive
traffic \citep{koopman2017safety}. Rulebooks and temporal-logic planners
specify priority among rules
\citep{censi2019rulebooks,wongpiromsarn2012receding,tumova2013least};
RSS-style models define responsibility and duties of care
\citep{shalev2017rss}; and reachability tools bound what agents can safely do
\citep{althoff2014online,althoff2016setbased}. These lines of work give modern
AV stacks powerful veto mechanisms. The less studied question is what should
happen after a veto when the scene is interactive and the violation is
repairable.

This paper studies the false-veto problem in rule-aware interactive driving. A
candidate merge, roundabout entry, or unsignalized-intersection crossing may
violate a hard gap margin under its current timing. A binary rule gate must
reject it. Yet human traffic often resolves exactly such cases through small,
lawful accommodations: the ego waits slightly, a non-priority vehicle yields
within its comfortable deceleration budget, or both parties make a minor
tactical edit. Treating the first negative margin as terminal discards these
recoverable interactions and biases the planner toward unnecessary conservatism.

Existing alternatives leave a critical gap. Ego-only trajectory repair can make
the ego wait or decelerate, but it cannot certify a bounded request to another
agent. Interaction-aware planners can predict that another driver will yield,
but then the safety argument depends on the correctness of a behavior model.
Hard-prune rulebooks give clear vetoes, but they do not produce a positive
witness explaining why a repaired maneuver is admissible. What is missing is a
runtime object that answers four questions together: which rule binds, which
finite edit repairs it, who owns the edit, and whether every requested
accommodation is affordable under right-of-way.

We call this object an \emph{interactive repair certificate}. A certificate is
not a predicted trajectory and not a learned confidence score. It is a compact
proof object that states the binding rule, the selected repair operators, their
owners and costs, the right-of-way-scaled request bounds, and the ego fallback
available if an elicited request is not observed. This formulation turns
interactive recovery into an auditable AI decision problem: learned or
rule-based planners may propose maneuvers, while the certificate layer decides
whether the proposed interaction has a bounded, attributable, normatively
admissible repair witness.

\carve\ instantiates this idea as a prediction-free repair layer over a finite
multi-owner tactical lattice. Its key mechanism is a cooperation envelope
\(\bj(s)=\beta(\pi_j)\amax(s)\), where \(\amax(s)\) is a kinematic
accommodation bound and \(\beta(\pi_j)\) scales that bound by the agent's
right-of-way status. Thus a non-priority agent may be asked for a small bounded
yield, while a priority agent receives no nonzero request. Because certificate
validity depends on the declared envelope and lattice rather than on a learned
response model, \carve\ can recover false vetoes without assuming another
driver will comply.

\begin{figure*}[t]
\centering
\includegraphics[width=0.94\textwidth]{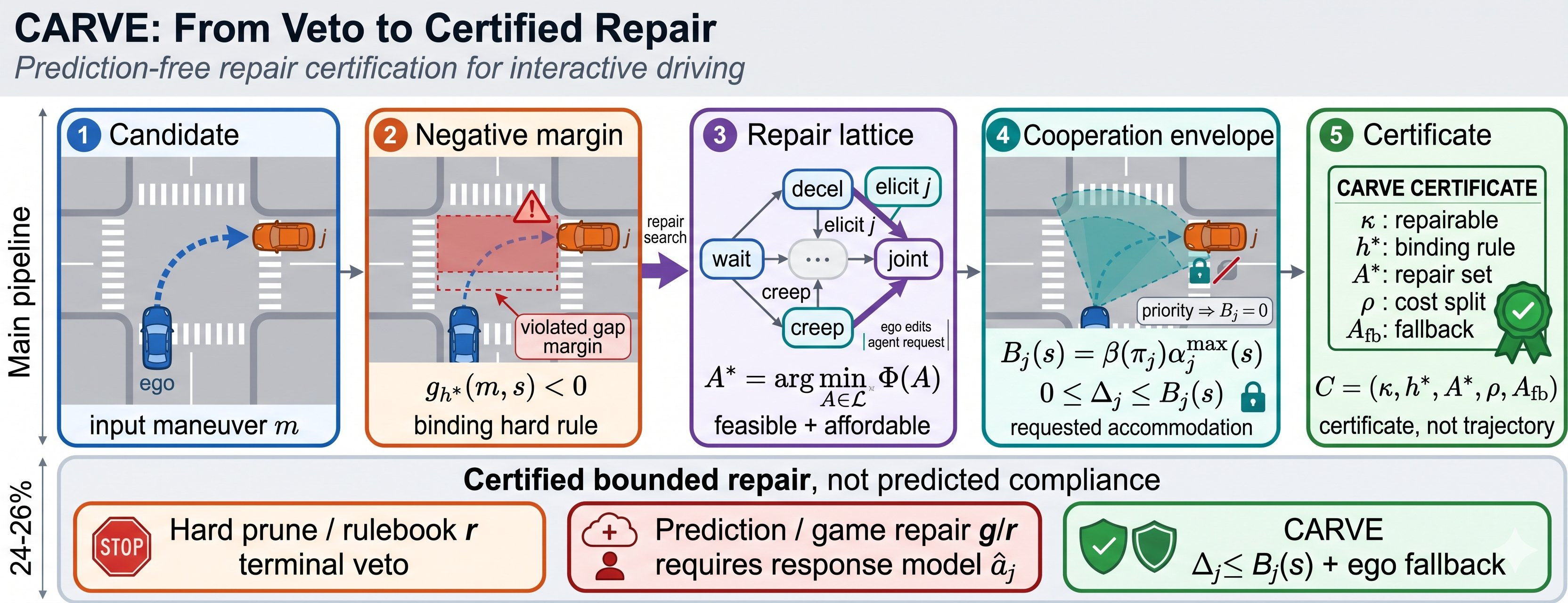}
\caption{Overview. \carve\ converts an initially infeasible interactive
candidate into a finite repair search and a certificate. Unlike
prediction-based repair, \carve\ certifies bounded requests and an ego fallback;
unlike hard-prune rulebooks, it can recover false vetoes through
right-of-way-affordable interaction.}
\label{fig:overview}
\end{figure*}

Our contributions are:
\begin{itemize}
\item We identify false-veto recovery as an interactive certification problem
rather than a trajectory-prediction problem, and formalize repair certificates
over a finite multi-owner operator lattice.
\item We introduce a right-of-way-scaled cooperation envelope
\(\bj=\beta(\pi_j)\amax\), which cleanly separates physical reachability from
normative admissibility and blocks requests to priority agents.
\item We give exact and greedy certificate procedures with soundness,
finite-lattice minimality, structural right-of-way respect, fallback
contingency, and multi-agent blame-consistency conditions.
\item We evaluate \carve\ on Lanelet2-geometry-grounded INTERACTION replay
episodes with ablations, negative stress tests, integrity checks, and
synthetic multi-agent blame-consistency stress, showing high false-veto
recovery without priority-agent false positives.
\end{itemize}

\section{Related Work}

\paragraph{Rule-aware planning and safety filters.}
Rulebooks, temporal logic, and minimum-violation synthesis reason about
prioritized specifications \citep{censi2019rulebooks,
wongpiromsarn2012receding,tumova2013least}. Control-barrier-function filters
and shielding methods provide runtime safety interventions in continuous
control or reinforcement learning \citep{ames2017cbf,alshiekh2018shielding}.
\carve\ is complementary: it operates at the tactical decision layer and emits
a human-readable multi-agent certificate, including repair ownership and
right-of-way affordability, rather than a continuous control correction alone.

\paragraph{Trajectory repair and reachability.}
Reachability analysis provides conservative envelopes for traffic participants
\citep{althoff2014online,althoff2016setbased}, and CommonRoad supplies
reproducible motion-planning benchmarks \citep{althoff2017commonroad}. The
closest ego-only repair work uses SMT and reachability to repair a violating
ego trajectory \citep{lin2024trafficrepair}. \carve\ instead searches a
discrete tactical lattice, includes agent-owned bounded accommodations, and
allocates repair burden across agents.

\paragraph{Prediction and interaction-aware planning.}
Motion forecasting has progressed from social recurrent models
\citep{alahi2016sociallstm,deo2018convsocial} to map-aware multimodal models
\citep{salzmann2020trajectron,chai2020multipath,gao2020vectornet}.
Game-theoretic and interaction-aware planners model how an ego action affects
other agents \citep{sadigh2016planning,fisac2019hierarchical,
schwarting2019social,fridovich2020ilq,hubmann2018uncertain}; interaction-aware
trajectory repair explicitly predicts responses \citep{wang2024interactionrepair}.
\carve\ does not replace these models. It audits a candidate maneuver and
certifies bounded requests without assuming that a response model is correct.

\paragraph{Recent AAAI decision-making context.}
Recent AAAI work studies regulation-aware driving decisions, search-based
vehicle planning, and safety-critical scenario generation
\citep{cai2026drivereg,nachkov2026dss,xu2025diffscene}. \carve\
is complementary: it does not generate trajectories or scenarios, but certifies
whether a proposed interactive maneuver has a bounded, attributable,
right-of-way-affordable repair witness.

\begin{table}[t]
\centering
\scriptsize
\begin{tabular}{lccccc}
\toprule
Family & Cert. & Bounded & RoW & Pred.-free & Blame \\
\midrule
Rulebook / hard gate & -- & -- & \(\checkmark\) & \(\checkmark\) & -- \\
Ego trajectory repair & -- & -- & -- & \(\checkmark\) & -- \\
Interaction prediction & -- & -- & -- & -- & -- \\
CBF / shielding & partial & -- & -- & \(\checkmark\) & -- \\
\textbf{\carve} & \(\checkmark\) & \(\checkmark\) & \(\checkmark\) & \(\checkmark\) & \(\checkmark\) \\
\bottomrule
\end{tabular}
\caption{Capability comparison. ``Bounded'' denotes explicit certification of
agent-owned requests inside a cooperation envelope.}
\label{tab:related}
\end{table}

\paragraph{Datasets and map grounding.}
Large datasets support perception and forecasting
\citep{caesar2020nuscenes,chang2019argoverse,sun2020waymo,ettinger2021waymo}.
Drone-based trajectory datasets provide clean kinematics
\citep{krajewski2018highd,bock2019ind}. INTERACTION targets adversarial and
cooperative driving with semantic maps \citep{zhan2019interaction}; Lanelet2
provides the map abstraction \citep{poggenhans2018lanelet2}. We use
INTERACTION as a replay test set only; \carve\ learns no parameters from it.

\section{Problem Formulation}

Let \(s\) be a scene with ego state, agents, map context, and right-of-way roles
\(\pi_j\). Let \(m\) be a candidate maneuver and
\(\rules=\{h_1,\ldots,h_L\}\) a prioritized hard-rule prefix. Each rule returns
a margin \(g_\ell(m,s)\); \(m\) is hard feasible iff all margins are
nonnegative. We use gap and time-to-conflict margins related to TTC and
car-following safety primitives \citep{minderhoud2001ttc,treiber2000idm,
kesting2010idm}.

\paragraph{Operator lattice.}
An operator is a tuple \(o=(\mathrm{owner},\mathrm{type},\Theta_o,
\cost_o,\gain_o)\). The owner is ego or a specific agent \(j\); \(\Theta_o\)
is a finite parameter grid; \(\cost_o(\theta)\ge 0\) is normalized effort; and
\(\gain_o(\theta)\ge 0\) is the margin gain on the binding rule. A repair set
\(\repair\) is a finite set of operator-parameter assignments. The lattice is
ordered by set inclusion over assignments; exact search finds a minimum-cost
feasible element of this finite lattice.

\paragraph{Objective and affordability.}
For a repair set \(\repair\), \carve\ minimizes
\begin{equation}
\phiobj(\repair)=
\sum_{(o,\theta)\in\repair_\ego}\cost_o(\theta)
+\sum_j w(\pi_j)\sum_{(o,\theta)\in\repair_j}\cost_o(\theta),
\end{equation}
where \(\cost_o\) is normalized operator effort and \(w(\pi_j)\) is the
responsibility weight used for blame ordering. Feasibility, however, is checked
in the units of the owner. Ego edits are charged to the normalized effort budget
\(B_\ego\). The two budgets are not compared to each other. If one or more
agent-owned requests target agent \(j\), their total positive speed-reduction
magnitude is \(\Req_j\) in m/s and must satisfy
\begin{equation}
0\le \Req_j \le \bj(s), \qquad
\bj(s)=\beta(\pi_j)\amax(s).
\end{equation}
Here \(\beta(\pi_j)\in[0,1]\) is dimensionless: priority, equal-duty, and
yielding agents use \(0,0.5,0.8\), respectively. This table is not learned or
fit to maximize recovery; it encodes an ordinal duty structure in which priority
agents receive zero request budget, equal-duty agents receive a conservative
partial envelope, and yielding agents receive a larger but still sub-reachability
envelope. The kinematic bound is the
closed-form speed-reduction inner bound
\begin{equation}
\amax(s)=\min(|a^{\min}_j|T,\|v_j\|_2),
\end{equation}
with horizon \(T=5\) s in the replay protocol. Thus \(B_j\) is in m/s, and the
affordability screen compares it to the requested accommodation magnitude. The
responsibility-weighted cost in \(\Phi\) is used for minimality and blame
ordering, not as the agent-side feasibility unit. This separation is why
tightening or loosening \(\beta\) changes recovery and CPA while structural
priority violations remain impossible when priority agents keep \(\beta=0\).

\paragraph{Elicitation semantics.}
\carve\ does not assume a communication channel. A certificate may be
implemented by V2X, a user-interface advisory, or a monitored hypothesis in the
planner. If the requested accommodation is not observed, the ego executes or
replans around the fallback. The certificate asserts admissibility of a
request, not compliance.

\paragraph{Certificate.}
A certificate is
\[
\cert=(\kappa,h^\star,\repair^\star,\cost_\ego,\{\cost_j\},\fallback),
\]
where \(\kappa\) is a category, \(h^\star\) is the binding rule,
\(\repair^\star\) is the selected repair, \(\cost_\ego,\{\cost_j\}\) are cost
allocations, and \(\fallback\) is an ego-only contingency when available.

\begin{figure*}[t]
\centering
\includegraphics[width=0.88\textwidth]{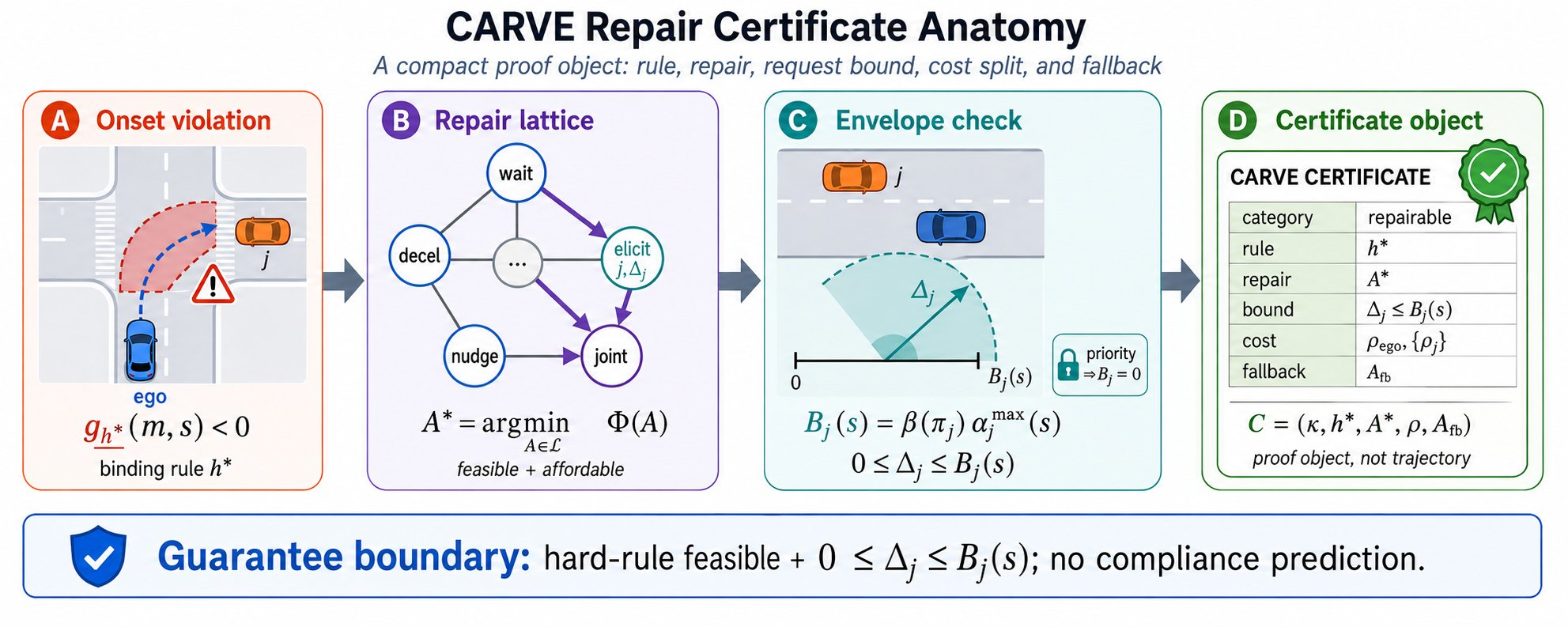}
\caption{Certificate anatomy. Blue elements denote ego-owned edits; teal
elements denote agent-owned accommodation requests. The guarantee boundary is
hard-rule feasibility plus \(0\le\Delta_j\le B_j(s)\), not prediction of
another driver's compliance.}
\label{fig:certificate}
\end{figure*}

\section{The CARVE Algorithm}

\begin{algorithm}[t]
\caption{\carve\ Decision Procedure}
\label{alg:carve}
\begin{algorithmic}[1]
\STATE \textbf{Input:} maneuver \(m\), scene \(s\), rules \(\rules\), operators \(\ops\), mode
\IF{\(g_\ell(m,s)\ge 0\) for all \(h_\ell\in\rules\)}
  \STATE \textbf{return} satisfied certificate with empty repair
\ENDIF
\STATE \(h^\star\leftarrow\) highest-priority rule with negative margin
\STATE \(d\leftarrow -g_{h^\star}(m,s)\); \(\pool\leftarrow\emptyset\)
\FOR{each operator \(o\in\ops\)}
  \IF{\(\exists\theta\in\Theta_o\) with positive gain and possible affordability}
    \STATE add feasible assignments \((o,\theta)\) to \(\pool\)
  \ENDIF
\ENDFOR
\IF{mode is exact}
  \STATE run branch-and-bound over nodes \((\repair,d_{\rm rem},\Phi)\)
  \STATE lower bound \(LB\leftarrow\) fractional cheapest-margin cover of \(d_{\rm rem}\)
  \STATE prune nodes with \(\Phi+LB\) no better than incumbent
\ELSE
  \STATE greedily add affordable assignment maximizing \(\gain/\cost\)
\ENDIF
\IF{no affordable feasible repair is found}
  \STATE \textbf{return} over-budget if feasibility appears only without budgets, else non-repairable
\ENDIF
\STATE compute best ego-only fallback \(\fallback\), if any
\STATE \textbf{return} category, \(\repair^\star\), cost split, and \(\fallback\)
\end{algorithmic}
\end{algorithm}

The exact solver uses an admissible fractional relaxation: it covers the
remaining negative margin by sorting unused assignments by cost per unit margin
gain and allowing fractional use. This can only underestimate the cost of a
discrete completion, so the standard branch-and-bound principle remains
admissible \citep{lawler1966branch}, analogous to classical heuristic search
with optimistic lower bounds \citep{pearl1984heuristics}. Greedy is the online
path; it is a fast anytime heuristic whose soundness is still checked by the
same feasibility and affordability predicates. Exact minimality is audited by
branch-and-bound. With pool size \(|P|\), grid size absorbed into \(P\), and
maximum selected depth \(D\), greedy is \(O(D|P|)\) after pool construction;
exact search is exponential in \(D\) in the worst case but finite and heavily
pruned by budget and bound checks.

\noindent\fbox{\begin{minipage}{0.94\columnwidth}
\textbf{What \carve\ certifies.}
Given declared hard-rule margins, a finite operator lattice, and declared
cooperation envelopes, an accepting certificate proves hard-rule feasibility
and affordability of the selected repair. It does not prove global continuous
optimality, infer legal ground truth from geometry proxies, or guarantee that
another driver will comply.
\end{minipage}}

\section{Guarantees}

All statements are with respect to the configured finite lattice, declared
rules, and declared envelopes. Proofs and assumption audits are in the
supplement. The assumptions are finite operator grids, deterministic declared
margins, monotone operator gains, admissible exact-search lower bounds,
unit-consistent envelopes, and exchangeability for the blame theorem.

\begin{theorem}[Certificate soundness]
If \carve\ returns an accepting certificate with repair \(\repair^\star\), then
\(m^{\repair^\star}\) satisfies every rule in \(\rules\), ego cost is within
\(B_\ego\), and every requested accommodation satisfies
\(0\le\Delta_j\le B_j(s)\).
\end{theorem}
This holds because acceptance is gated by recomputing all hard margins and the
affordability predicate on the selected set.

\begin{theorem}[Structural right-of-way respect]
If agent \(j\) is labeled priority, then an accepting certificate contains no
positive accommodation request owned by \(j\).
\end{theorem}
The mechanism is structural: \(\beta(\pi_j)=0\) implies \(B_j=0\), so any
positive request fails the affordability screen.

\begin{theorem}[Finite-lattice minimality]
Exact \carve\ returns a minimum-\(\Phi\) affordable repair over the finite
operator lattice, or reports that no affordable feasible repair exists.
\end{theorem}
The lower bound never overestimates the remaining completion cost; therefore a
pruned node cannot contain a better incumbent.

\begin{theorem}[Fallback contingency boundary]
For elicited or joint certificates with an ego fallback, the ego has an
executable contingency if the requested accommodation is not observed.
\end{theorem}
The guarantee is a runtime boundary, not a compliance prediction.

\begin{theorem}[Blame consistency]
For distinct-duty agents with exchangeable accommodation operators and
sufficient capacity, exact \carve\ cannot allocate more raw accommodation
magnitude to a lower-duty agent than to a higher-duty agent.
\end{theorem}
Here exchangeable means that equal raw speed-reduction magnitudes have the same
margin-channel effect for the compared agents. Equal-duty agents are
intentionally unordered. The theorem orders raw accommodation magnitude, not
weighted cost. The result follows from a swap argument: moving burden from a
lower-duty, higher-weight agent to a higher-duty, lower-weight agent lowers
\(\Phi\), contradicting exact minimality.

\section{Evaluation Protocol}

\paragraph{Episode mining.}
We mine replay episodes using INTERACTION trajectories and Lanelet2 map
geometry. When regulatory right-of-way metadata is available, we record it
explicitly; otherwise, we assign a geometry-derived proxy label and report that
source separately. Inclusion requires: (i) an ego-agent conflict near a
Lanelet2 conflict point; (ii) an ego-unilateral hard gate veto at onset; (iii)
observed human resolution later in the window; and (iv) non-fragmented tracks
with unambiguous intent. No learning or tuning is performed on INTERACTION.

\begin{table}[t]
\centering
\scriptsize
\begin{tabular}{@{}p{0.46\columnwidth}rp{0.24\columnwidth}@{}}
\toprule
Audit stage & Count & Purpose \\
\midrule
Lanelet2-grounded replay set & 589 & final evaluation set \\
Initial hard-gate vetoes & 589 & no satisfied-onset cases \\
Ego-only repair accepted & 211 & floor comparator \\
Human-resolved ego-only vetoes & 378 & FVRR denominator \\
\carve\ accepted / refused & 581 / 8 & certificate outcome \\
\bottomrule
\end{tabular}
\caption{Auditable replay funnel after deterministic filters. Raw-data
regeneration, thresholds, and schema are in the supplement and code artifact.}
\label{tab:mining}
\end{table}

\paragraph{Baselines.}
HardPrune is the terminal-veto baseline. EgoOnly-Greedy and EgoOnly-Exact use
the same rules, costs, parameter grids, and search procedures as \carve, with
only agent-owned operators removed. UniversalYield-UpperBound is a diagnostic,
not a deployable method: it sets all agents to willing yielders and shows the
unconstrained cooperative ceiling after removing normative right-of-way
constraints. AlphaOnly-CARVE removes \(\beta\) from \(B_j\), and NoElicit
removes agent-owned operators.

\paragraph{Metrics.}
Accept is computed over all 589 replay episodes. FVRR is false-veto recovery
over the 378 human-resolved ego-only vetoes. CPA is an envelope diagnostic, not
a prediction-accuracy metric: for eligible elicited or joint certificates with a
measurable realized response, CPA is one when the observed accommodation is
inside \(B_j\). BCR is computed over eligible certificates with nonzero
agent-side cost allocations. Fallback is reported over elicited or joint
certificates where the finite lattice also contains an ego-only contingency.
RHA is repair-human agreement; it is diagnostic only because \carve\ searches
for minimal certificates rather than imitating human repair magnitudes.

\section{Results}

\begin{table*}[t]
\centering
\scriptsize
\begin{tabular}{lrrrrrrrr}
\toprule
Method & Accept & FVRR & RoW-resp. & Prio. FP & CPA(\(B_j\)) & BCR & Fallback & p50/p99 ms \\
\midrule
HardPrune & 0.00 & n/a & 100.00 & 0 & n/a & n/a & n/a & n/a \\
EgoOnly-Greedy & 35.82 & n/a & 100.00 & 0 & n/a & n/a & n/a & n/a \\
EgoOnly-Exact & 35.82 & n/a & 100.00 & 0 & n/a & n/a & n/a & n/a \\
UnivYield-UB (diag.) & 100.00 & n/a & n/a & n/a & n/a & n/a & n/a & n/a \\
AlphaOnly-CARVE & 100.00 & 100.00 & 97.62 & 14 & 83.36 & 100.00 & 35.82 & n/a \\
\textbf{\carve-Greedy} & \textbf{98.64} & \textbf{97.88} & \textbf{100.00} & \textbf{0} & 66.20 & \textbf{100.00} & 35.54 & 1.76/2.99 \\
\carve-Exact audit & 98.64 & 97.88 & 100.00 & 0 & 66.49 & 100.00 & 33.93 & 2.67/11.61 \\
\bottomrule
\end{tabular}
\caption{Main results on 589 Lanelet2-geometry-grounded INTERACTION replay
episodes. Values are percentages except false-positive counts and latency.
``n/a'' means the metric is undefined for that comparator; the upper-bound row
is an unconstrained diagnostic, not a deployable baseline. For \carve-Greedy,
FVRR is 370/378, RoW-respect is 589/589, BCR is 574/574, and negative-stress
veto is 400/400. Latency was measured on a 4-core/8-thread laptop CPU.}
\label{tab:main}
\end{table*}

Table~\ref{tab:main} gives the central contrast. Ego-only repair, with
identical rules, costs, grids, and search but with agent-owned operators removed,
accepts only 35.82\% of the mined interactive conflicts. \carve-Greedy accepts
98.64\% and recovers 370/378 false vetoes, showing that the missing capacity is
lawful multi-owner cooperation rather than a weaker search configuration. This
recovery does not come from ignoring normative priority: RoW-respect is 589/589,
priority false positives are zero, and BCR is 574/574. The diagnostic upper
bound accepts every case only by allowing universal yielding; AlphaOnly also
reaches 100\% FVRR but produces 14 priority-agent false positives.

\begin{figure}[!t]
\centering
\includegraphics[width=\columnwidth]{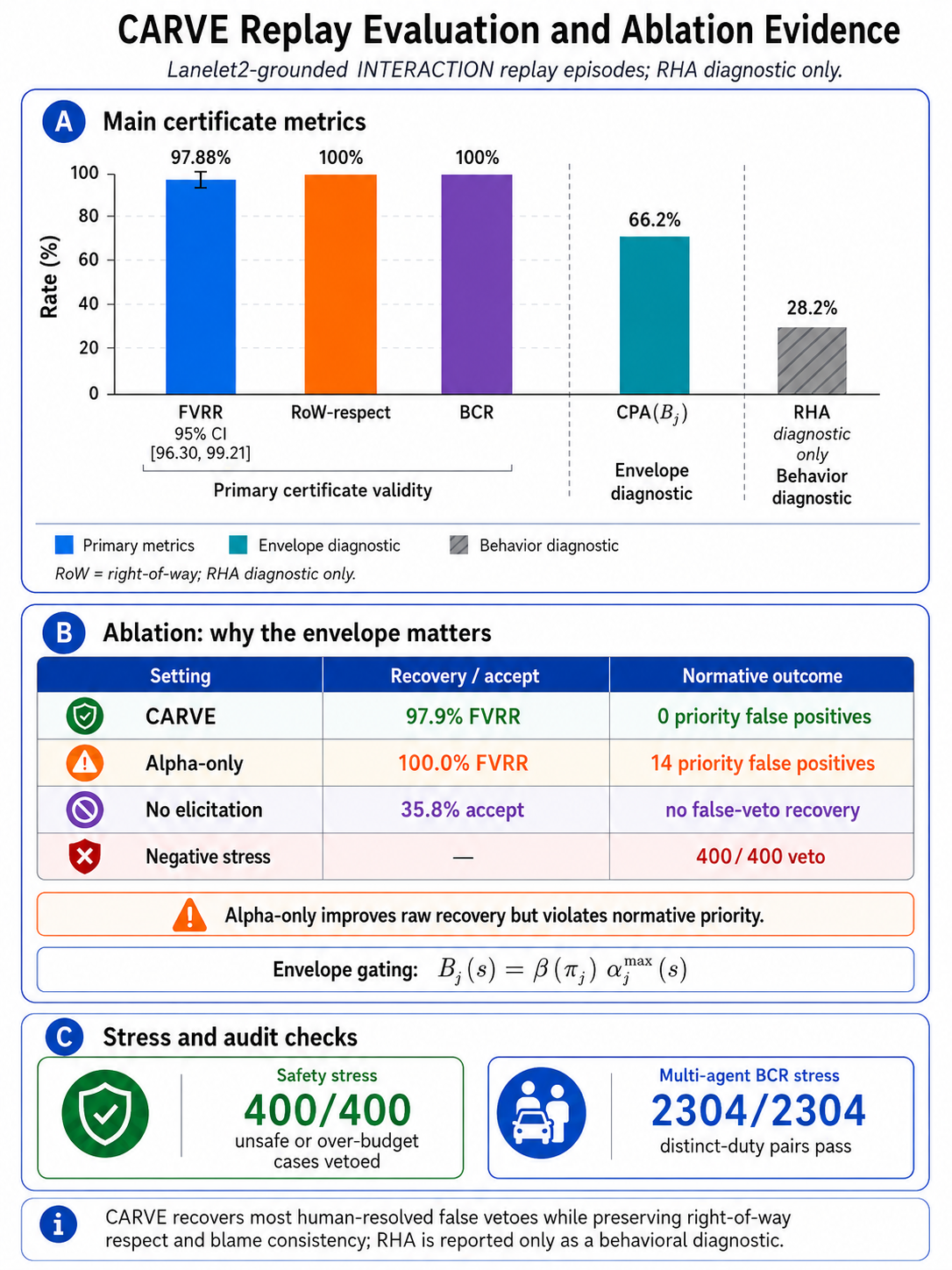}
\caption{Replay evaluation and ablation evidence. The main certificate
metrics are FVRR, RoW-respect, and BCR; CPA(\(B_j\)) is an envelope diagnostic;
RHA is a behavior diagnostic. Alpha-only improves raw recovery but violates
priority; negative stress and synthetic BCR stress audit safety and
responsibility properties.}
\label{fig:results}
\end{figure}

\paragraph{Sensitivity to hand-designed parameters.}
The operator coefficients are declared protocol parameters, not learned. To
test whether the result is a tuning artifact, we sweep gain, ego-cost,
agent-cost, TTC-threshold, and \(\beta\)-table families. Across all non-priority
envelope sweeps, RoW-respect remains 100.00\% and priority false positives
remain zero. A strict \(\beta\) table lowers FVRR to 95.50\% and CPA to 48.57\%;
a permissive table keeps FVRR at 97.88\% and raises CPA to 79.34\%; AlphaOnly
raises CPA further but creates 14 priority false positives. Thus the key
invariant comes from the structural priority envelope, not a tuned coefficient.

\paragraph{Failure analysis.}
The eight unrecovered false vetoes are categorized as repairable only outside
the declared budgets. They are therefore not silent failures: \carve\ refuses
to certify repairs that require priority yielding or exceed \(B_j\). CPA
failures similarly indicate that observed human accommodation sometimes exceeds
the conservative normative envelope, not that the certificate predicted the
wrong behavior. A higher CPA can be obtained by loosening the envelope, but that
is not a win when it admits priority requests. RHA separates type from magnitude:
broad repair type matches 359/581 cases, while full type-plus-magnitude RHA is
164/581, consistent with humans often choosing larger, later, or more
comfortable maneuvers than a minimal certificate. The default certificate
categories are 7 ego-only, 563 elicited, 11 joint, and 8 over-budget refusals.
Fallback coverage is not required for every accepted certificate; for 204/574
elicited or joint certificates an ego-only contingency exists, otherwise runtime
commitment must monitor and recertify.

\paragraph{Row-source and multi-agent audits.}
The replay set spans 11 INTERACTION locations. We keep row-source labels
separate because many labels are geometry-derived proxies, not explicit
regulatory right-of-way.

\begin{table}[t]
\centering
\scriptsize
\begin{tabular}{@{}lrl@{}}
\toprule
Row source & N & Claim status \\
\midrule
all-way stop & 143 & stop-type metadata \\
geometry-arrival proxy & 198 & proxy label \\
geometry-equal proxy & 204 & proxy label \\
explicit regulatory RoW & 1 & explicit metadata \\
partial regulatory RoW & 43 & partial metadata \\
\bottomrule
\end{tabular}
\caption{Right-of-way source stratification. All episodes are
Lanelet2-geometry grounded; proxy rows are not claimed as explicit regulatory
labels.}
\label{tab:rowsource}
\end{table}

Naturalistic
multi-agent evidence is modest: 45 two-agent and 2 three-agent conflict scenes
yield 24 eligible real pairwise BCR checks. We therefore separately use a
synthetic distinct-duty stress set, which passes 648/648 BCR scenes and
2304/2304 pairwise checks. The synthetic set validates the formal property; it
is not presented as naturalistic scale.

\paragraph{Negative and integrity checks.}
Negative stress consists of 200 unrepairable collision cases and 200
priority-overbudget cases where the only feasible repair would require a
nonzero request to a priority agent with \(\beta=0\). \carve\ vetoes 400/400.
Integrity scripts replay saved decisions and find zero main-set or
negative-set failures. The package includes 17 unit tests covering
affordability units, right-of-way invariants, branch-and-bound admissibility,
operator monotonicity, and Lanelet2 parser invariants.

\FloatBarrier
\section{Discussion and Limitations}

\carve\ is a certification layer, not a full AV stack. It can wrap learned
behavior planners, trajectory generators, or rule-based candidate generators by
auditing proposed maneuvers. This supports trustworthy and explainable AI:
black-box components may propose, while \carve\ returns an auditable decision
object naming the binding rule, responsible party, request bound, and fallback.

The guarantees are relative to a finite tactical lattice, declared margins, and
conservative envelopes. They do not imply global continuous optimality,
closed-loop safety under unmodeled rules, or another driver's compliance.
Open-loop replay evaluates certificate logic against observed human
resolutions; this is the appropriate first test for the certificate logic, but
it does not measure closed-loop social feedback. Closed-loop validation with
reactive agents in CommonRoad or CARLA is future work. Only a small subset of
the replay episodes contains explicit regulatory RoW metadata; the rest are
reported as all-way-stop metadata or geometry-derived proxies. The current
lattice targets one to three conflict agents; dense scenes require hierarchical
conflict clustering, and repairs outside the finite tactical ontology are out of
scope.

The evaluation does not claim that \carve\ outperforms all interaction-aware
planners. UniversalYield is an upper bound, not an official external
implementation. The comparison isolates a different question: whether an
initially vetoed maneuver admits a bounded, attributable,
right-of-way-respecting repair certificate.

\section{Conclusion}

We introduced \carve, a certificate-generating repair layer for interactive
driving. \carve\ turns infeasibility into a minimal, affordable, and
blame-consistent repair object without predicting another agent's response. On
INTERACTION replay episodes it recovers most human-resolved false vetoes while
preserving structural right-of-way respect and vetoing unsafe stress cases.
This makes repairability a first-class, auditable decision object: not merely
whether a maneuver is allowed, but whose bounded action would make it allowed
and under what normative envelope.

\bibliography{references}

\end{document}